\documentclass[10pt,twocolumn,letterpaper]{article}

\usepackage[pagenumbers]{cvpr} 

\usepackage[dvipsnames]{xcolor}

\definecolor{cvprblue}{rgb}{0.21,0.49,0.74}
\usepackage[pagebackref,breaklinks,colorlinks,citecolor=cvprblue]{hyperref}
\usepackage{float}
\usepackage{wrapfig}
\usepackage{multirow}
\usepackage{color, colortbl}
\usepackage{graphicx}
\usepackage{soul}
\usepackage[ruled,linesnumbered]{algorithm2e}

\usepackage{amsmath,amsfonts,bm}

\def\eqref#1{equation~\ref{#1}}

\def\1{\bm{1}}

\def\ve{{\bm{e}}}
\def\vf{{\bm{f}}}

\def\vn{{\bm{n}}}

\def\vx{{\bm{x}}}

\def\mM{{\bm{M}}}

\def\mV{{\bm{V}}}

\DeclareMathAlphabet{\mathsfit}{\encodingdefault}{\sfdefault}{m}{sl}
\SetMathAlphabet{\mathsfit}{bold}{\encodingdefault}{\sfdefault}{bx}{n}
\newcommand{\tens}[1]{\bm{\mathsfit{#1}}}

\def\tI{{\tens{I}}}

\def\gE{{\mathcal{E}}}

\def\gG{{\mathcal{G}}}

\def\gV{{\mathcal{V}}}

\newcommand{\R}{\mathbb{R}}

\usepackage{soul}
\usepackage{marvosym}

\def\paperID{4724} 
\def\confName{CVPR}
\def\confYear{2024}

\title{MemoNav: Working Memory Model for Visual Navigation}

\author{
  Hongxin Li$^{1,2}$\quad 
  Zeyu Wang$^{1,2}$ \quad
  Xu Yang$^{1,2}$ \quad 
  Yuran Yang$^{5}$ \quad
  Shuqi Mei$^{5}$\quad 
  Zhaoxiang Zhang$^{1,2,3,4}$\textsuperscript{\Letter} \quad
  \\ 
$^1$University of Chinese Academy of Sciences (UCAS) \\
$^2$State Key Laboratory of Multimodal Artificial Intelligence Systems, CASIA \\
$^3$Center for Artificial Intelligence and Robotics, HKISI, CAS \\
$^4$Shanghai Artificial Intelligence Laboratory \quad
$^5$Tencent Maps, Tencent \\
  \small{Code: \url{https://github.com/ZJULiHongxin/MemoNav}}
}

\begin{document}
\maketitle
\begin{abstract}
Image-goal navigation is a challenging task that requires an agent to navigate to a goal indicated by an image in unfamiliar environments. Existing methods utilizing diverse scene memories suffer from inefficient exploration since they use all historical observations for decision-making without considering the goal-relevant fraction. To address this limitation, we present MemoNav, a novel memory model for image-goal navigation, which utilizes a working memory-inspired pipeline to improve navigation performance. Specifically, we employ three types of navigation memory. The node features on a map are stored in the short-term memory (STM), as these features are dynamically updated. A forgetting module then retains the informative STM fraction to increase efficiency. We also introduce long-term memory (LTM) to learn global scene representations by progressively aggregating STM features. Subsequently, a graph attention module encodes the retained STM and the LTM to generate working memory (WM) which contains the scene features essential for efficient navigation. The synergy among these three memory types boosts navigation performance by enabling the agent to learn and leverage goal-relevant scene features within a topological map. Our evaluation on multi-goal tasks demonstrates that MemoNav significantly outperforms previous methods across all difficulty levels in both Gibson and Matterport3D scenes. Qualitative results further illustrate that MemoNav plans more efficient routes.
\end{abstract}    
\section{Introduction}
\label{sec:intro}
Image-goal navigation (ImageNav) is an attractive embodied AI task where an agent is guided toward a destination indicated by an image within unfamiliar environments. This task has garnered significant attention recently, owing to its promising applications in enabling robots to navigate open-world scenarios.

\begin{figure}[t]
  \centering
   \includegraphics[width=0.95\linewidth]{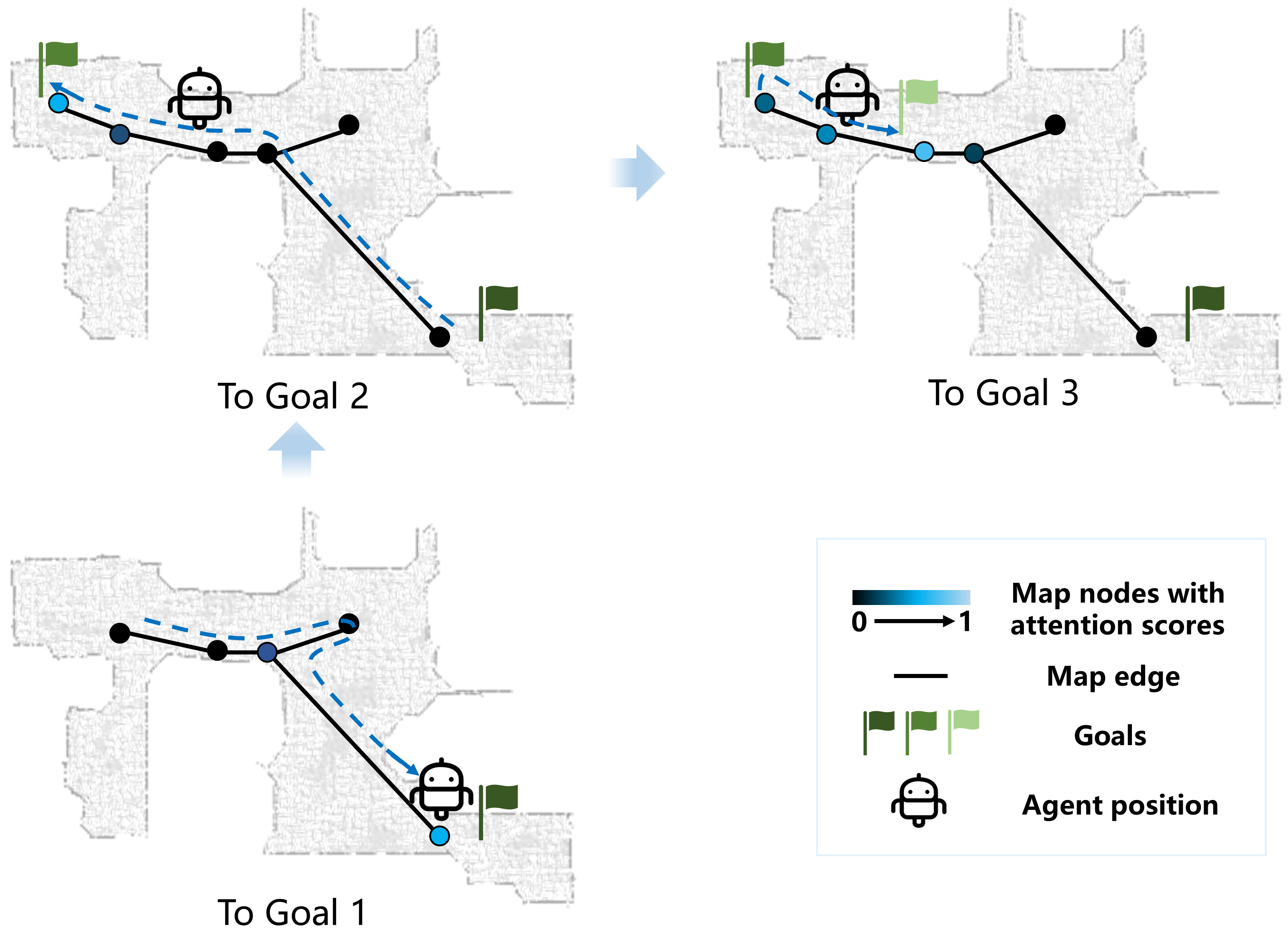}

   \caption{\textbf{A brief example of MemoNav.} MemoNav calculates attention scores for each node on the topological map and then excludes the nodes with low scores (the black nodes in the figure) during decision-making. This design helps our agent focus more on goal-relevant scene features, boosting multi-goal visual navigation performance.}
   \label{fig:MemoNav_effect}
\end{figure}

Central to ImageNav is scene memory, which serves as a repository of crucial historical information for decision-making in unseen environments~\cite{savinov2018semi}. During navigation, this memory typically stores both scene features and the agent's navigation history~\cite{kwon2021visual}, thereby enhancing navigation by mitigating the challenges of partial observability~\cite{parisotto2018neural}. In literature, various memory mechanisms have been introduced for ImageNav, which can be classified into three categories according to memory structure: (a) metric map-based methods~\cite{chaplot2020learning, chen2018learning} that reconstruct local top-down maps and aggregate them into a global map, (b) stacked memory-based methods~\cite{pashevich2021episodic, Mezghani2021MemoryAugmentedRL, fang2019scene} that stack the past observations chronologically, and (c) topological map-based methods~\cite{savinov2018semi, kwon2021visual, chaplot2020neural, beeching2020learning, kim2023topological} that store sparse landmark features in graph nodes. Notably, topological map-based methods leverage the sparsity of topological maps, demonstrating impressive performance in ImageNav.

Nevertheless, existing topological map-based methods still suffer from two major limitations: 
(a) Unawareness of useful nodes. These methods typically use all node features for generating actions without considering the contribution of each node, thus being easily misled by redundant nodes that are uninformative of the goal.
(b) Local representation. Each node feature only represents a small area in a large scene, limiting the agent's capacity to learn a higher-level semantic and geometric representation of the entire scene.

To address these limitations, we present a novel ImageNav method named MemoNav (refer to~\cref{fig:MemoNav_effect}), which draws inspiration from the classical concept of working memory in cognitive neuroscience~\cite{cowan2008differences} and loosely aligns with the working memory model in human navigation~\cite{blacker2017keeping}.

MemoNav learns three types of scene representations: \textbf{Short-term memory} (STM) represents the local and transient features of nodes in a topological map. \textbf{Long-term memory} (LTM) represents a global node that acquires a scene-level representation by continuously aggregating STM. \textbf{Working memory} (WM) learns goal-relevant features about 3D scenes and is used by a policy network to generate actions. The WM is formed by encoding the informative fraction of the STM and the LTM.

\begin{figure*}[th]
    \centering
    \includegraphics[width=0.9\linewidth]{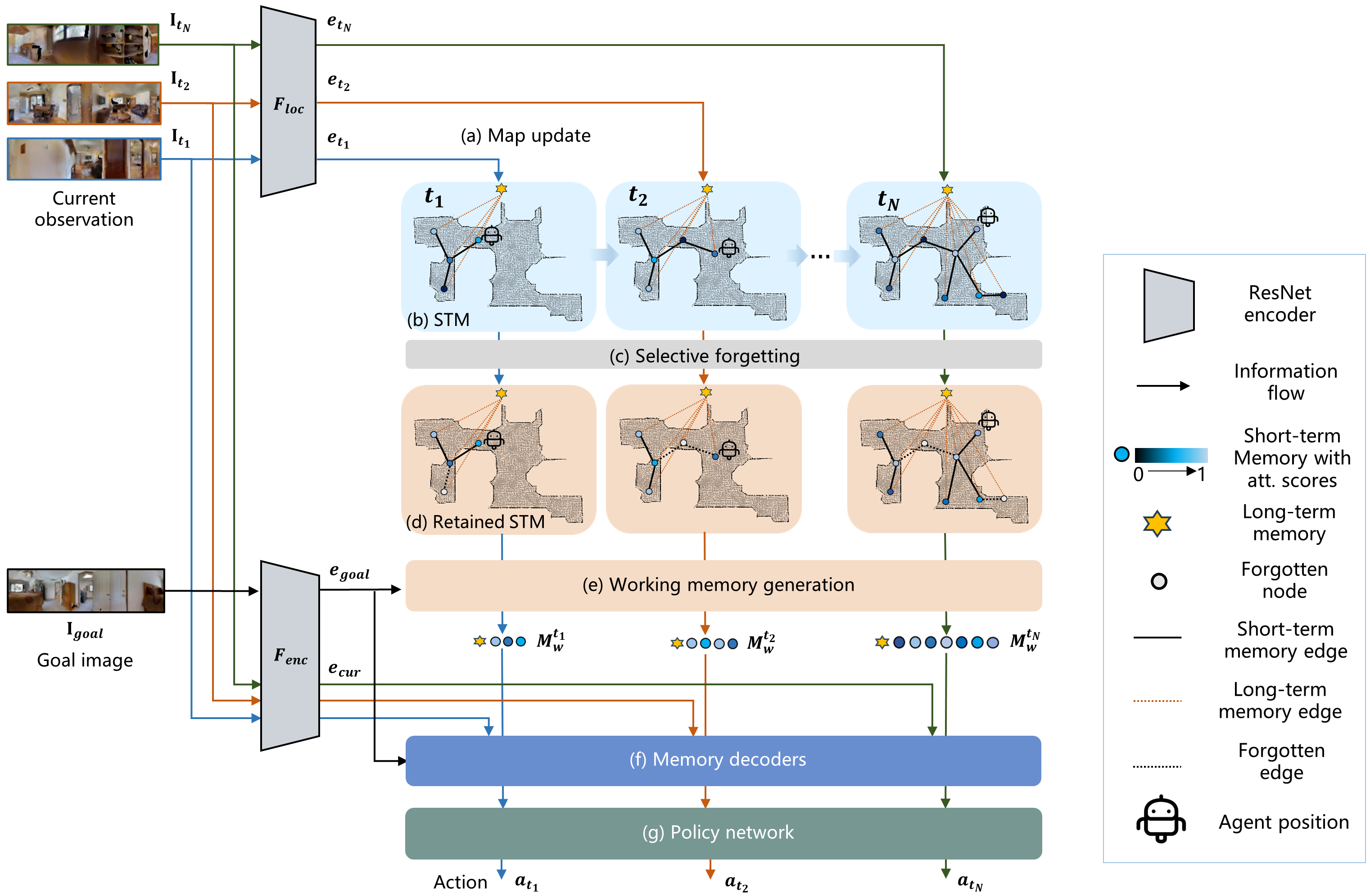}
    \caption{\textbf{Overview of MemoNav.} (a) The memory update module builds a topological map using $\displaystyle \ve_t$, the embedding of the current image $\displaystyle \tI_t$. (b) The node features in the map constitute the STM while a global node that links to each node acts as the LTM. (c) The forgetting module temporarily excludes a fraction of STM whose attention scores rank below a threshold $\displaystyle p$. (d) The retained STM and the LTM are concatenated and then encoded by (e) a graph attention module to generate the WM $\displaystyle \mM_w^t$. (f) The WM is decoded by two Transformer decoders (details in~\cref{fig:detailed_memonav}). (g) Lastly, the output of the decoding process is input to a policy network to generate navigation actions.}
    \label{fig:model_overview}
\end{figure*}

Based on the above three representations, MemoNav navigation pipeline (\cref{fig:model_overview}) contains five steps: (1) STM generation. The map updating module stores landmark features on the map as STM. (2) Selective forgetting. A forgetting module incorporates goal-relevant STM into WM by temporarily removing nodes with attention scores ranking below a predefined threshold. After this process, the navigation pipeline excludes the forgotten nodes in subsequent time steps. (3) LTM generation. To assist STM, a global node is added to the map as LTM. This node links to all map nodes and continuously aggregates their features at each time step. (4) WM generation. A graph attention module encodes the retained STM and LTM to generate WM. The WM combines goal-relevant information from STM with scene-level features from LTM, enhancing the agent's ability to use informative scene representations for improved navigation. (5) Action generation. Two Transformer decoders use the embeddings of the goal image and the current observation to decode the WM. The decoded features are then used to generate navigation actions.

Consequently, with the synergy of the three representations, MemoNav outperforms state-of-the-art methods in the Gibson scenes~\cite{xia2018gibson}, enjoying substantial improvements on multi-goal navigation tasks. Comparison in the Matterport3D scenes~\cite{chang2017matterport3d} also highlights MemoNav's superiority.

The main contributions of this paper are as follows:
\begin{itemize}
\item We propose MemoNav, which learns three types of scene representations (STM, LTM, and WM) to improve navigation performance in the ImageNav task.
\item We use a forgetting module to retain informative STM, thereby reducing redundancy in the map and improving navigation efficiency. We also introduce a global node as the LTM, connecting to all STM nodes and providing a global scene-level perspective to the agent.
\item We adopt a graph attention module to generate WM from the retained STM and the LTM. This module utilizes adaptive weighting to generate effective WM used for challenging multi-goal navigation.
\item The experimental results demonstrate that our method outperforms existing methods on both 1-goal and multi-goal tasks across two popular scene datasets.
\end{itemize}

\section{Related Work}
\label{sec:related work}

\noindent\textbf{ImageNav methods}. Since an early attempt~\cite{zhu2017target} to train agents in a simulator for ImageNav, rapid progress has been made on this task  \cite{beeching2020learning,chen2021topological,hahn2021no,kwon2021visual,du2021curious,wasserman2022last, majumdar2022zson, al2022zero,yadav2023offline}. Several methods have utilized topological scene representations for visual navigation, of which SPTM~\cite{savinov2018semi} is an early work. NTS~\cite{chaplot2020neural}, VGM~\cite{kwon2021visual}, and TSGM~\cite{kim2023topological} incrementally build a topological map during navigation and generalize to unseen environments without exploring the scenes in advance. These methods utilize all features in the map, while our MemoNav flexibly utilizes the informative fraction of these features. Another line of work~\cite{yadav2023offline, majumdar2022ssl} has introduced self-supervised learning to enhance the scene representations, achieving a promising navigation success rate. In contrast, we enhance the scene representations using the proposed LTM that aggregates the agent's local observation features.

\noindent\textbf{Memory models for reinforcement learning}. Several studies \cite{Qiao2023BraininspiredIR, Zou2022TowardsAN, ritter2021rapid, lampinen2021towards, sukhbaatar2021not, loynd2020working} draw inspiration from memory mechanisms of the human brain and design reinforcement learning models for reasoning over long time horizons. Ritter et al. \cite{ritter2021rapid} proposed an episodic memory storing state transitions for navigation tasks. Lampinen et al. \cite{lampinen2021towards} presented hierarchical attention memory as a form of ``mental time-travel'' \cite{tulving1985memory}, which means recovering goal-oriented information from past experiences. Unlike this method, our model retains such information via a novel forgetting module. Expire-span \cite{sukhbaatar2021not} predicts life spans for each memory fragment and permanently deletes expired ones. Our model is different from this work in that we restore forgotten memory if the agent returns to visited places. \cite{loynd2020working} shares a similar idea but just solves simple 2D grid-world tasks~\cite{babyai} and its memory capacity is fixed. In contrast, our method employs an adaptive working memory to tackle more complicated long-horizon navigation tasks.
\section{Background}
\label{sec:background}
\subsection{Task Definition}
\label{sec:bg:task def}

The objective of ImageNav is to learn a policy $\pi$ to reach a goal, given an image $\tI_{goal}$ that contains a view of the goal and a series of observations $\{\tI_t \}$ captured during the navigation. At the beginning of navigation, the agent receives an RGB image $\tI_{goal}$ of the goal. At each time step, the agent captures an RGB-D panoramic image $\tI_{t}$ of the current location and generates a navigational action. Following~\cite{kwon2021visual}, any additional sensory data (e.g., GPS and IMU) are not available.

\subsection{Brief Review of Visual Graph Memory}
\label{sec:bg:VGM}

Our MemoNav is primarily based on Visual Graph Memory (VGM)~\cite{kwon2021visual}, which is briefly introduced below.
VGM incrementally builds a topological map $\gG = (\gV, \gE)$ from the agent's past observations where $\gV$ and $\gE$ denote nodes and edges, respectively. The node features $\mV \in \R^{d \times N_{t}}$ are derived from observations by a pretrained encoder $ \mathcal{F}_{loc}$ where $d$ denotes the feature dimension and $N_{t}$ the number of nodes at time $t$.

VGM uses a graph convolutional network (GCN) to encode the topological map into a memory representation $\mM=\operatorname{GCN}(\mV)$. Before encoding, VGM obtains the goal embedding $\ve_{goal}=\mathcal{F}_{enc}(\tI_{goal})$ and fuses each node feature with this embedding through a linear layer.

The encoded memory is then decoded by two Transformer~\cite{vaswani2017attention} decoders, $\mathcal D_{cur}$ and $\mathcal D_{goal}$. $\mathcal D_{cur}$ takes the current observation embedding $\ve_{cur}=\mathcal{F}_{enc}(\tI_{cur})$ as the query and the feature vectors of the encoded memory $\mM$ as the keys and values, generating a feature vector $\vf_{cur}$. Similarly, $\mathcal D_{goal}$ takes the goal embedding $\ve_{goal}$ as the query and generates $\vf_{goal}$. Lastly, a LSTM-based policy network takes as input the concatenation of $\vf_{cur}$, $\vf_{goal}$ and $\ve_{cur}$ to output an action distribution.
\section{Method}
\label{sec: components}

MemoNav integrates three principal components: the forgetting module, long-term memory generation, and working memory generation. We illustrate the pipeline of the MemoNav in~\cref{fig:model_overview} and detail these components in this section. We also briefly discuss the connection between our method and working memory studies~\cite{cowan2008differences, baddeley2012working} in the appendix.

\subsection{Selective Forgetting Module}
\label{sec: forgetting mechanism}
MemoNav continually adds nodes to its topological map during exploration. We denote these nodes as short-term memory (STM)\footnote{The two terminologies "Nodes" and "STM" will be used interchangeably in the following sections.} as they are dynamically substituted with new ones when the agent revisits corresponding areas.

Our pilot studies indicated that not all STM equally contribute to reaching goals. We visualize the attention scores for the STM calculated in the memory decoder $\mathcal D_{goal}$ (\cref{fig:MemoNav_effect} and \cref{fig:supp:traj_demos} in the appendix). The figures show that high scores are assigned to nodes leading to goals while little attention is paid to remote ones. This phenomenon suggests that it is more efficient to use the goal-relevant fraction of scene memory. According to this finding, we devise a forgetting module that enables the agent to forget uninformative experiences. Here, ``forgetting'' means that STM with attention scores lower than a threshold are temporarily excluded from the navigation pipeline. This means of forgetting via attention is also evidenced by research~\cite{fukuda2009human} suggesting that optimal working memory performance depends on focusing on task-relevant information.

The forgetting module retains a fraction of STM according to the attention scores $\displaystyle \{\alpha_{i}\}_{i=1}^{N_t}$ in $\mathcal D_{goal}$. These scores reflect the extent to which the goal embedding $\displaystyle \ve_{goal}$ attends to each STM feature. After $\mathcal D_{goal}$ finishes decoding, the agent temporarily ``forgets'' a fraction of nodes whose scores rank below a predefined percentage $\displaystyle p$, meaning these nodes are disconnected from their neighbors and excluded from the navigation pipeline in subsequent steps. If the agent returns to a forgotten node, this node will be re-added to the map and processed by the pipeline again. In multi-goal tasks, once a goal is reached, all forgotten nodes will be restored for potential usefulness in locating the next goal. The forgetting module operates in a plug-and-play manner, which means it is not activated during training but switched on during evaluation and deployment (refer to~\cref{fig:detailed_memonav} for details). $\displaystyle p$ is set as 20\% as we empirically find that this suits most tasks. With this module, the agent can selectively retain the informative STM, while avoiding misleading experiences.

\subsection{Long-Term Memory Generation}
\label{sec: Global node}

In addition to STM, the information in the long-term memory (LTM) also forms part of working memory (WM)~\cite{ericsson1995long}. Inspired by ETC~\cite{ainslie2020etc} and LongFormer~\cite{beltagy2020longformer}, we add a zero-initialized trainable global node $\displaystyle \vn_{global} \in \R^d$ to the topological map, representing the LTM (the orange star in~\cref{fig:model_overview}), which connects to all nodes in the map and aggregates the STM features at each time step.  Unlike RecBERT~\cite{hong2021vln}, which uses a recurrent state token to encode visual-linguistic clues, our LTM aggregates the agent's past observations by continuously fusing the STM through memory encoding (the encoder is described in the next subsection).

LTM offers two key benefits: it learns a scene-level feature and facilitates feature fusion. A recent study~\cite{ramakrishnan2022environment} suggests that embodied agents benefit from higher-level environment representations to mitigate partial observability from limited field-of-view sensors. From this viewpoint, the LTM stores a high-level scene representation by aggregating local node features. Moreover, the LTM facilitates feature fusion, especially useful when the topological map is segmented into isolated sub-graphs due to the removal of forgotten nodes. By connecting to every node, the LTM acts as a bypath aiding in feature fusion across these sub-graphs.

\subsection{Working Memory Generation}
The third type of scene representation WM learns goal-relevant features for action generation. To learn adaptive WM, we utilize a graph attention module GATv2~\cite{brody2022how} to encode the retained STM and the LTM, capitalizing on GATv2's effectiveness in scenarios where nodes have varying neighbor importance. GATv2 adaptively assigns weights to neighboring nodes based on their features, instead of relying on a static Laplacian matrix. This design is suitable for generating WM, especially in multi-goal tasks since the STM features related to a path leading to the goal should obtain high weights while those of irrelevant places should receive lower weights. In addition, as STM features are updated with new features upon revisiting nodes, GATv2's adaptive weighting is particularly suitable for aggregating the STM features into the WM. The WM generation process is formulated as follows:
\begin{equation}
\displaystyle \mM_w = \{\mV^{\prime}, \vn_{global}^t\}=\operatorname{GATv2}(\{\mV, \vn_{global}^{t-1}\})
\label{eq:WM}
\end{equation}
where $\displaystyle \mM_w$ represents the generated WM, and $\displaystyle \mV^{\prime}$ the encoded STM. $\displaystyle  \{\cdot, \cdot\}$ denotes that the LTM (a vector) is appended to the retained STM (a sequence of vectors). Note that the time step superscript of $\displaystyle \vn_{global}$ means that the LTM is recurrent through time.

After GATv2 encoding, the WM aggregates the goal-relevant information from retained STM as well as the scene-level representation from the LTM. Lastly, the decoders $\mathcal D_{cur}$ and $\mathcal D_{goal}$ take $\displaystyle \mM_w$ as keys and values, generating $\displaystyle \vf_{cur}$ and $\displaystyle \vf_{goal}$, which are further used to generate actions.
\section{Experiments}

\begin{figure}[t]
  \centering
   \includegraphics[width=0.85\linewidth]{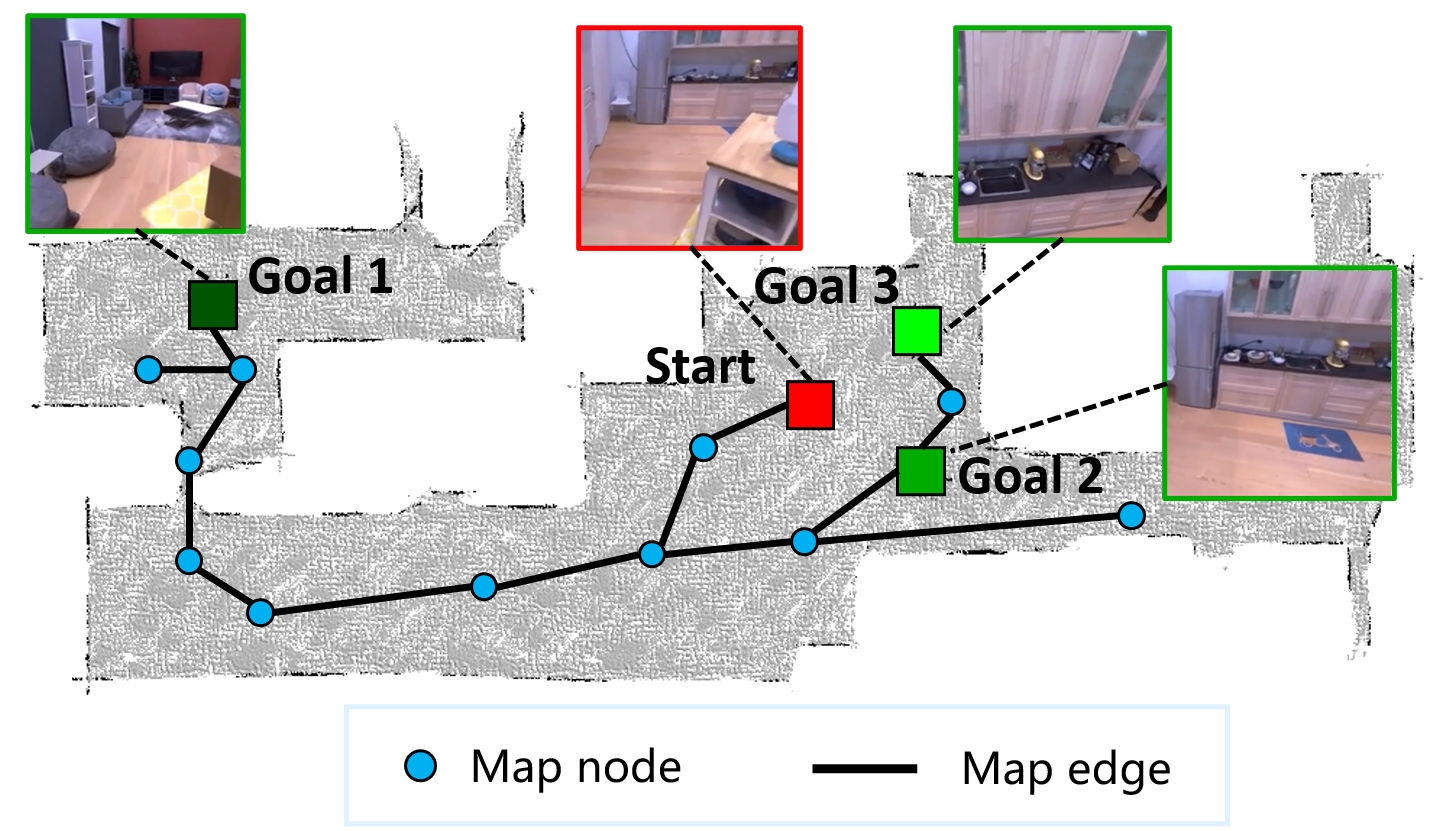}
   \caption{\textbf{An example episode for multi-goal tasks in Gibson.} The agent is tasked with navigating to multiple sequential goals.}
   \label{fig:task_comparison}
\end{figure}
\begin{table*}[t]
\normalsize

\centering
\begin{tabular}{@{}ccccccccccc@{}}
\toprule
\multirow{2}{*}{Scene} &  \multirow{2}{*}{Methods}  & \multicolumn{2}{c}{\textbf{1-goal}} & \multicolumn{2}{c}{\textbf{2-goal}} & \multicolumn{2}{c}{\textbf{3-goal}} & \multicolumn{2}{c}{\textbf{4-goal}} \\ \cmidrule(lr){3-4} \cmidrule(lr){5-6} \cmidrule(lr){7-8}  \cmidrule(lr){9-10}
 & &  \textbf{SR} & \textbf{SPL} & \textbf{PR} & \textbf{PPL} & \textbf{PR} & \textbf{PPL} & \textbf{PR} & \textbf{PPL} \\ \midrule
\multirow{5}{*}{G}

                  & ANS~\cite{chaplot2020learning} & $30.0$ &  $11.0$ & - & - & - & - & - & - \\
                  & NTS~\cite{chaplot2020neural} & $43.0$ & $26.0$ & - & - & - & - & - & - \\
                  & CNNLSTM~\cite{zhu2017target} & $53.1$ & $39.2$ & $31.5$ & $10.6$ & $18.0$ & $2.8$ & $12.4$ & $1.6$\\
                  
                  & TSGM~\cite{kim2023topological} & $70.3$ & $50.0$ & 
                  $27.8$ & $16.1$ & $17.4$ & $\mathbf{10.4}$ & $13.4$ & $4.6$ \\
                  & VGM~\cite{kwon2021visual} & $70.0$ & $55.4$ &  
                  $42.9$ & $17.1$ & $29.5$ & $7.0$ & $21.5$ & $4.1$ \\   
                  & MemoNav (ours)  & $\mathbf{74.7}$ & $\mathbf{57.9}$ 
                  & $\mathbf{50.8}$ & $\mathbf{20.1}$ & $\mathbf{38.0}$ & $9.0$ & $\mathbf{28.9}$ & $\mathbf{5.1}$ 
                  \\ \midrule

\multirow{3}{*}{M}
                   & CNNLSTM~\cite{zhu2017target} & $16.2$ & $9.8$ & $10.8$ & $2.6$ & $7.7$ & $1.4$ & - & -\\
                   & TSGM~\cite{kim2023topological} & $24.0$ & $14.6$ &  
                   $13.5$ & $\mathbf{6.2}$ & $7.8$ & $\mathbf{3.8}$ & - & - \\
                   & VGM~\cite{kwon2021visual} & $25.1$ & $\mathbf{16.6}$ & $16.7$ & $5.0$ & $11.8$ & $2.5$ & - & - \\
                   & MemoNav (ours)  & $\mathbf{26.1}$ & $16.3$ & $\mathbf{19.5}$ & $5.6$ & $\mathbf{13.6}$ & $2.9$ & - & - \\
\bottomrule
\end{tabular}
\caption{\textbf{Comparison between MemoNav and previous methods.} The evaluation results in Gibson (G) and Matterport3D (M) scenes demonstrate that MemoNav outperforms previous methods across all difficulty levels. Note that the 1-goal evaluation in Gibson uses 1007 hard episodes following~\cite{kwon2021visual} while the multi-goal evaluation uses our collected episodes. \textbf{SR}: success rate (\%), \textbf{SPL}: success weighted by path length (\%), \textbf{PR}: progress (\%), \textbf{PPL}: progress weighted by path length (\%).}
\label{tab: multi goal}
\end{table*}
\subsection{Datasets}
\label{sec: exp details}

All experiments are conducted in the Habitat~\cite{savva2019habitat} simulator with the Gibson \cite{xia2018gibson} and Matterpot3D \cite{chang2017matterport3d} scene dataset, adopting an action space consistent with VGM~\cite{kwon2021visual}. 

\noindent\textbf{1-goal evaluation.}
In Gibson, we use 72 scenes for training and a public dataset \cite{Mezghani2021MemoryAugmentedRL} comprising 14 unseen scenes for evaluation. Following the setting in VGM \cite{kwon2021visual}, 1007 out of 1400 1-goal episodes\footnote{The 1-goal difficulty level here denotes the hard level in this public test dataset} from this public dataset are used for evaluation, while we still use the full set of 1400 episodes for ablation studies of MemoNav.

\noindent\textbf{Multi-goal dataset.}
Multi-goal evaluation, which requires an agent to navigate to an ordered sequence of goals, is more suitable for evaluating memory models used for navigation. By enabling the agent to return to visited places we can test whether memory models help the agent plan efficient paths. If not, the agent will probably waste its time re-exploring the scene or traveling randomly. However, recent ImageNav methods seldom conduct multi-goal evaluations. To further investigate the efficacy of MemoNav, we follow MultiON~\cite{wani2020multion} to compile 700-episode multi-goal test datasets in the Gibson scenes (see~\cref{fig:task_comparison} for an example).

\begin{figure*}[th]
    \centering
    \includegraphics[width=0.84\linewidth]{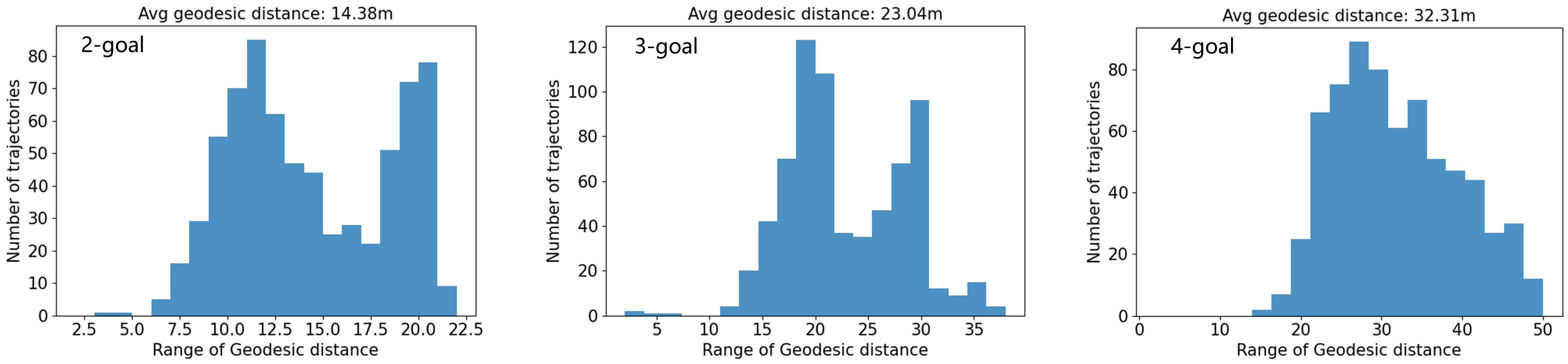}
    \caption{Histograms of geodesic distances for the multi-goal test datasets. As we set a distance limit for goals and discard invalid trajectories, a scene may own its prominent distance range, leading to the nonuniform histograms.}
    \label{fig:geohist}
\end{figure*}

We follow five rules to set sequential goals for each episode: (1) No obstacles appear near each goal. (2) The distance between two successive goals is no more than 10 meters. (3) All goals are placed on the same layer. (4) All goals are reachable from each other. (5) The final goal is placed near a certain previous one. Please refer to~\cref{fig:geohist} for dataset statistics.

In Matterpot3D, we sample 1008 episodes per difficulty level from the multi-goal test datasets used in Multi-ON \cite{wani2020multion}. The difficulty of an episode is indicated by the number of goals. All methods are trained on the Gibson 1-goal dataset and tested across varying difficulty levels.

\noindent\textbf{Evaluation Metrics}. In 1-goal tasks, the success rate (\textbf{SR}) and success weighted by path length (\textbf{SPL}) \cite{anderson2018evaluation} are used. In a multi-goal task, two metrics are borrowed from \cite{wani2020multion}: The progress (\textbf{PR}) is the fraction of goals successfully reached, equal to the \textbf{SR} for 1-goal tasks; Progress weighted by path length (\textbf{PPL}) indicates navigation efficiency and is defined as $\displaystyle PPL=\frac{1}{E} \sum_{i=1}^{E} Progress_{i} \frac{l_{i}}{\max \left(p_{i}, l_{i}\right)}$, where $\displaystyle E$ is the total number of test episodes, $\displaystyle l_i$ and $\displaystyle p_i$ are the shortest path distance to the final goal via midway ones, and the actual path length taken by the agent, respectively. The objective of each goal is to stop within 1 meter of the goal location and each episode is allowed 500 steps.

\begin{table*}[t]

\normalsize 
\centering

\begin{tabular}{@{}cccccccccccc@{}}
\toprule
\multirow{2}{*}{} & \multicolumn{3}{c}{\textbf{Components}} & \multicolumn{2}{c}{\textbf{1-goal}} & \multicolumn{2}{c}{\textbf{2-goal}} & \multicolumn{2}{c}{\textbf{3-goal}} & \multicolumn{2}{c}{\textbf{4-goal}} \\ \cmidrule(lr){2-4} \cmidrule(lr){5-6} \cmidrule(lr){7-8} \cmidrule(lr){9-10}  \cmidrule(lr){11-12}

  & \textbf{Forget} & \textbf{LTM} & \textbf{WM} & \textbf{SR} & \textbf{SPL} & \textbf{PR} & \textbf{PPL} & \textbf{PR} & \textbf{PPL} & \textbf{PR} & \textbf{PPL} \\ \midrule

1 &          &         &           & $52.1$ & $46.7$ & $42.9$ & $17.1$ & $29.5$ & $7.0$ & $21.5$ & $4.1$ \\

2 & \checkmark         &         &   & $55.1$ & $46.1$ & $44.9$ & $17.5$ & $29.4$ & $6.5$ & $21.5$ & $4.2$ \\

3 &          & \checkmark   &           & $58.9$ & $49.7$ & $43.8$ & $17.8$ & $29.6$ & $6.9$ & $25.1$ & $4.0$ \\

4 &  \checkmark        & \checkmark  &   & $60.6$ & $49.9$ & $48.1$ & $19.5$ & $37.5$ & $\mathbf{9.1}$ & $28.8$ & $4.9$ \\

5 &  & \checkmark  & \checkmark &$61.1$  & $48.9$ & $47.6$ & $17.8$ & $33.7$ & $7.9$ & $27.4$ & $5.0$ \\

6 & \checkmark  & \checkmark & \checkmark  & $\mathbf{62.4}$  & $\mathbf{50.7}$ & $\mathbf{50.8}$ & $\mathbf{20.1}$ & $\mathbf{38.0}$ & $9.0$ & $\mathbf{28.9}$ & $\mathbf{5.1}$ \\
\bottomrule
\end{tabular}

\caption{\textbf{Network component ablation results}. This ablation uses the entire 1400 1-goal episodes collected by~\cite{Mezghani2021MemoryAugmentedRL} and our collected multi-goal evaluation datasets. Row 1 is the baseline model VGM~\cite{kwon2021visual}, and row 6 is our full model. The table shows that when applied separately, the forgetting module and the LTM both improves performance and that the combination of these two components brings larger gains. Moreover, the synergy among the three components leads to the best performance. (\textbf{Forget}: Forgetting nodes with attention scores below 20\%, \textbf{LTM}: Using the LTM to continuously fuse the STM, \textbf{WM}: Using GATv2 to learn adaptive working memory)}
\label{tab: component ablation}
\end{table*}
\subsection{Compared Methods and Training Details}

We compare with the following methods which adopt various memory types:
\textbf{CNNLSTM}~\cite{zhu2017target} uses no maps but a hidden vector as implicit memory. \textbf{ANS}~\cite{chaplot2020learning} is a metric map-based model for ImageNav.
\textbf{NTS}~\cite{chaplot2020neural} incrementally builds a topological map without pre-exploring and adopts a hierarchical navigation strategy. \textbf{VGM}~\cite{kwon2021visual} is the baseline for MemoNav and has been elaborated in~\cref{sec:bg:VGM}. \textbf{TSGM}~\cite{kwon2021visual} associates a topological map with detected objects to use more semantic scene information.

We follow the training pipeline in~\cite{kwon2021visual} to reproduce CNNLSTM and train our MemoNav. These methods are first trained via imitation learning and then further fine-tuned with proximal policy optimization (PPO)~\cite{schulman2017proximal} (details in the appendix). The evaluation results for ANS and NTS are borrowed from the VGM paper~\cite{kwon2021visual}\footnote{ANS is designed for exploration while NTS is not open-sourced, so it is not straightforward to reproduce them for multi-goal tasks.}. All methods are equipped with a panoramic camera.

\subsection{Quantitative Results}
\noindent\textbf{Comparison on Gibson.}
\cref{tab: multi goal} shows that the MemoNav outperforms all compared methods in SR across all difficulty levels. Notably, CNNLSTM exhibits the poorest performance as its limited memory provides insufficient scene information. MemoNav also outperforms the metric map-based method ANS which requires pre-built maps. Compared with VGM, our model exhibits a noticeable performance gain, especially on the multi-goal tasks, enhancing SR by \textbf{7.9\%}. \textbf{8.5\%}, and \textbf{7.4\%} on the 2, 3, and 4-goal tasks respectively, while relying on less scene memory.


\begin{table*}[t]
\normalsize 
\centering
\begin{tabular}{@{}cccccccccc@{}}
\toprule
\multirow{2}{*}{} & \multirow{2}{*}{\textbf{Variants}} & \multicolumn{2}{c}{\textbf{1-goal}} & \multicolumn{2}{c}{\textbf{2-goal}} & \multicolumn{2}{c}{\textbf{3-goal}} & \multicolumn{2}{c}{\textbf{4-goal}} \\
\cmidrule(lr){3-4} \cmidrule(lr){5-6} \cmidrule(lr){7-8} \cmidrule(lr){9-10}
 &   & \textbf{SR} & \textbf{SPL} & \textbf{PR} & \textbf{PPL} & \textbf{PR} & \textbf{PPL} & \textbf{PR} & \textbf{PPL} \\ \midrule

1 & MemoNav                      & $62.4$ & $50.7$ & $50.8$ & $20.1$ & $38.0$ & $9.0$ & $28.9$ & $5.1$ \\

2 & LTM excluded  & $60.5$ & $49.1$ & $46.3$ & $16.3$ & $34.2$ & $8.0$ & $26.8$ & $4.9$ \\

3 & Random replacing  & $61.0$ & $47.9$ & $45.9$ & $17.4$ & $34.9$ & $8.2$ & $27.0$ & $5.0$ \\

4 & Averaging STM as LTM    & $59.1$ & $49.2$ & $46.1$ & $18.6$ & $37.4$ & $8.3$ & $28.4$ & $4.8$ \\

\bottomrule
\end{tabular}
\caption{\textbf{Ablation study of LTM}. Row 1 is our default model. Row 2 is a variant using the LTM to aggregates STM but does not incorporate the LTM into WM. Row 3 shows the impact of replacing the LTM with a random STM feature. Row 4 shows the result of replacing the LTM with averaged STM. The three variants disable the LTM, all leading to performance drops.}

\label{tab:LTM ablation}
\end{table*}
\noindent\textbf{Comparison on Matterport3D}.
We extend our evaluation to the Matterport3D scenes to assess the models' ability to generalize to different scene types. \cref{tab: multi goal} shows that our method achieves consistent performance improvements on this unseen scene dataset. Compared with VGM, MemoNav demonstrates higher SR/PR across the three difficulty levels. TSGM obtains slightly better PPL on multi-goal tasks probably because it uses more object-level clues to locate the goal. As the introduction of object semantics in TSGM is orthogonal to our contributions, we believe adding these semantics to MemoNav will lead to higher performance.

Overall, these results demonstrate that MemoNav benefits from the informative scene memory and the high-level scene representation contained in the WM, obtaining high success rates.

\subsection{Ablation Studies and Analysis}
\label{sec: ablation}

We conduct ablation studies in the Gibson scenes to analyze the impact of each proposed component.

\noindent\textbf{Performance gain from each proposed component.}
We assess the three key components outlined in~\cref{sec: components} and present the results in~\cref{tab: component ablation}. We can see that applying the forgetting module achieves improvements in the SR/PR (row 2 vs. row 1) and that the LTM also brings noticeable gains in SR/PR over the baseline (row 3 vs. row 1). Notably, the combined use of the forgetting module and LTM results in even larger increases (row 4 vs. row 1). More importantly, the synergy among the three components increases the SR/PR by \textbf{10.3\%}, \textbf{7.9\%}, \textbf{8.5\%}, and \textbf{7.4\%} at the 1, 2, 3, and 4-goal levels, respectively (row 6 vs. row 1). Comparing rows 6 and 5, we see that adding the forgetting module to MemoNav leads to notable increases, especially on multi-goal tasks. The qualitative evaluation in~\cref{fig: ablation vis} also shows that MemoNav without forgetting tends to take redundant steps, which also justifies the efficacy of the forgetting module. Overall, these results underscore the effectiveness of our components in addressing long-horizon navigation tasks with multiple sequential goals.

\noindent\textbf{The Critical Role of LTM.}
To study the effect of the LTM proposed in~\cref{sec: Global node}, we implement three distinct ways of disenabling the function of the LTM and show the results in~\cref{tab:LTM ablation}. The first ablation (row 2) excludes the LTM from the WM, preventing the high-level scene feature from being utilized by the downstream policy network. This modification worsens the performance across all difficulty levels. The second ablation replaces the LTM feature with a randomly selected STM feature each time the WM is generated according to~\cref{eq:WM}, resulting in inferior multi-goal navigation performances.

To justify that aggregating STM features into the LTM using adaptive weighting helps to learn a better high-level scene representation, we replace the LTM feature with an average of all STM features (row 4). After replacing, the LTM no longer learns a scene-level representation but still facilitates message passing among STM. Although this variant exhibits a decline in performance, the decrease is less pronounced for the more challenging 3 and 4-goal tasks. We hypothesize that this is because the simply averaged STM features act as a rudimentary scene-level feature but possess a weak representation capability compared to a learned LTM feature.

Overall, these findings confirm the LTM's vital role in learning scene-level features that are crucial for improving navigation performance.

\begin{figure*}[th]
    \centering
    \includegraphics[width=0.83\linewidth]{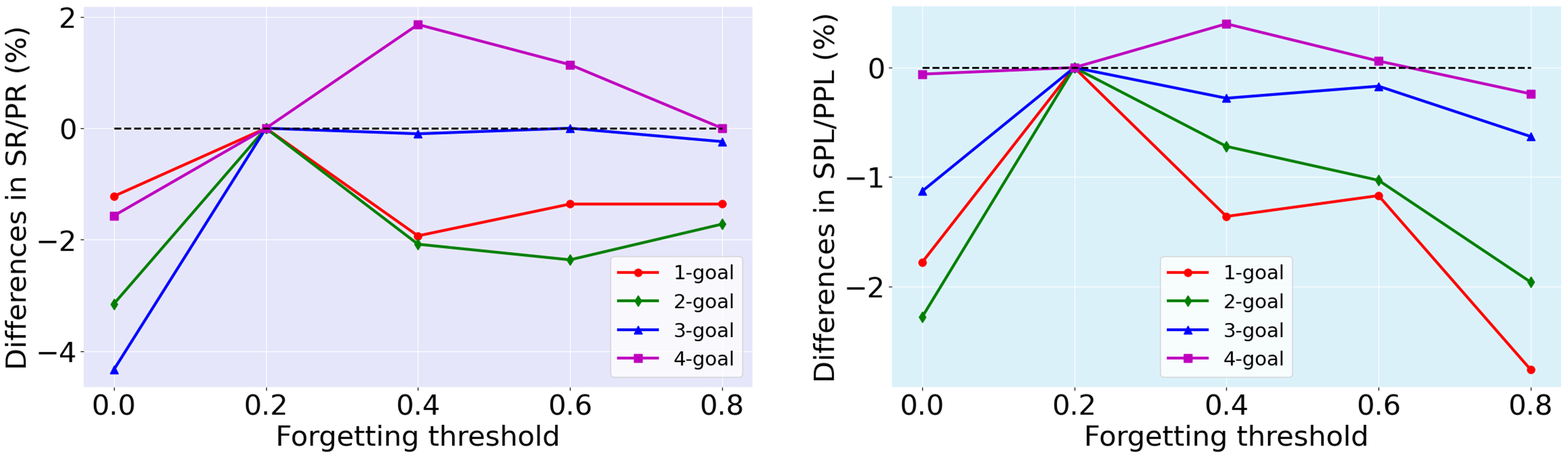}
    \caption{\textbf{Navigation performance versus forgetting threshold $\displaystyle p$ in the Gibson scenes.} MemoNav achieves the best performance on easier tasks with a lower $\displaystyle p$ but a higher $\displaystyle p$ is more beneficial for harder tasks. Moreover, MemoNav maintains high SR/PR with just 20\% of STM on the 3-goal tasks and enjoys a higher $\displaystyle p$ on the 4-goal tasks.}
    \label{fig:forgetting_th}
\end{figure*}

\begin{figure*}
  \centering
  \includegraphics[width=0.84\linewidth]{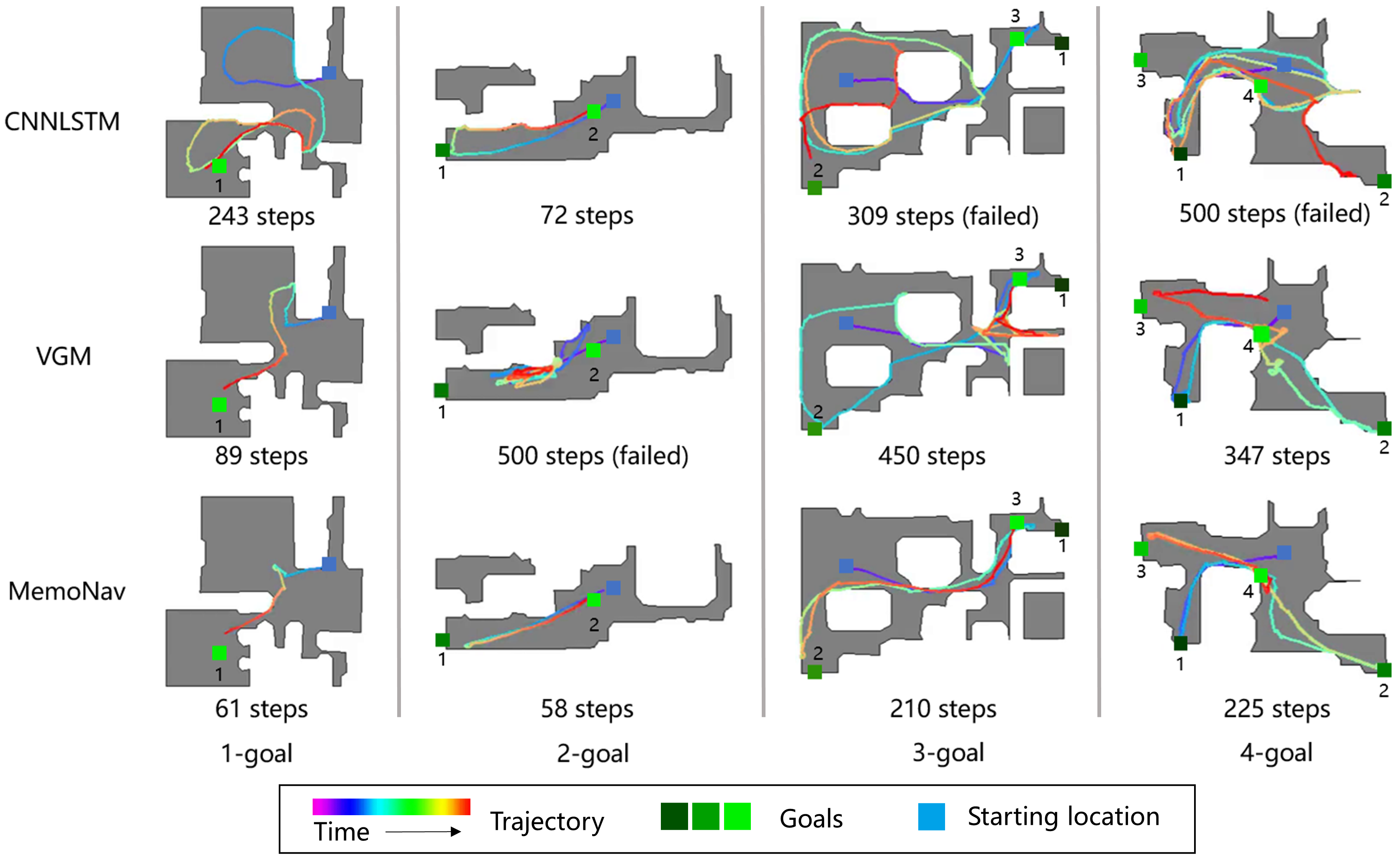}
  \caption{\textbf{Visualization of example episodes from a top-down view.} We compare CNNLSTM, VGM, and MemoNav at four difficulty levels in the Gibson scenes. Our MemoNav plans more efficient paths compared to the other two methods. For instance, in the 3-goal example, MemoNav quickly reaches the third goal which is located at an explored area. Best viewed in color.}  
  \label{fig:exp:traj_comp}
\end{figure*}

\noindent\textbf{Correlation between navigation performance and forgetting threshold}.
We evaluate our model with different forgetting thresholds $\displaystyle p$, as defined in~\cref{sec: forgetting mechanism}. The results are shown in~\cref{fig:forgetting_th}. For clarity, the figure shows the performance differences between our full model and the variants. The figure in general shows that MemoNav achieves the best performance on easier tasks with a lower $\displaystyle p$ but requires a higher $\displaystyle p$ for peak performance on harder tasks. Specifically, increasing $\displaystyle p$ from 0.0 (no forgetting) to 0.8 leads to initial improvements in SR/PR and SPL/PPL for 1 and 2-goal tasks, followed by a decline. As the agent seldom revisits explored areas in these easier tasks, using a larger $\displaystyle p$ to remove too much STM leads to a situation where the agent forgets what it has explored and takes more steps to re-explore the scene.

In contrast, for the more demanding 3 and 4-goal tasks, a higher $\displaystyle p$ is more helpful. For instance, at $\displaystyle p=0.8$, MemoNav exhibits only slight drops in SR/PR and SPL/PPL on 3-goal tasks, indicating that our agent is able to maintain a high success rate with just 20\% of STM. Notably, MemoNav achieves top performance in 4-goal tasks at $\displaystyle p=0.4$, suggesting that a larger portion of the STM can be forgotten when conducting longer-horizon navigation tasks that require an agent to frequently revisit explored places. The rationale behind this phenomenon is that if an agent has explored most of the scene after finishing several goals, it is supposed to plan shorter paths to subsequent goals by utilizing a small, goal-relevant fraction of STM.

We have also conducted an in-depth analysis of this phenomenon by plotting distributions of distances from retained/forgotten STM to each goal, as shown in~\cref{fig:supp:disthist} of the appendix. This analysis demonstrates that a substantial portion of retained nodes align closely with the shortest paths, indicating that MemoNav leverages a higher $\displaystyle p$ to focus more on the regions along short paths in multi-goal navigation scenarios.

\subsection{Qualitative Comparison}
To qualitatively assess MemoNav, we show example episodes of CNNLSTM, VGM, and MemoNav in the Gibson scenes in~\cref{fig:exp:traj_comp}. MemoNav's efficacy is evident in its shorter and smoother trajectories. Conversely, CNNLSTM exhibits extensive exploration steps and often fails in complex scenes due to its simplistic hidden states which struggle to encapsulate an effective scene memory. VGM frequently navigates in redundant circles, particularly in narrow pathways. For instance, as depicted in the second and third columns, MemoNav adeptly avoids the bottlenecks that entrap VGM, exemplifying MemoNav's capabilities of planning efficient routes.\par

\subsection{Visualization of MemoNav}
\begin{figure*}[th]
    \centering
    \includegraphics[width=0.95\linewidth]{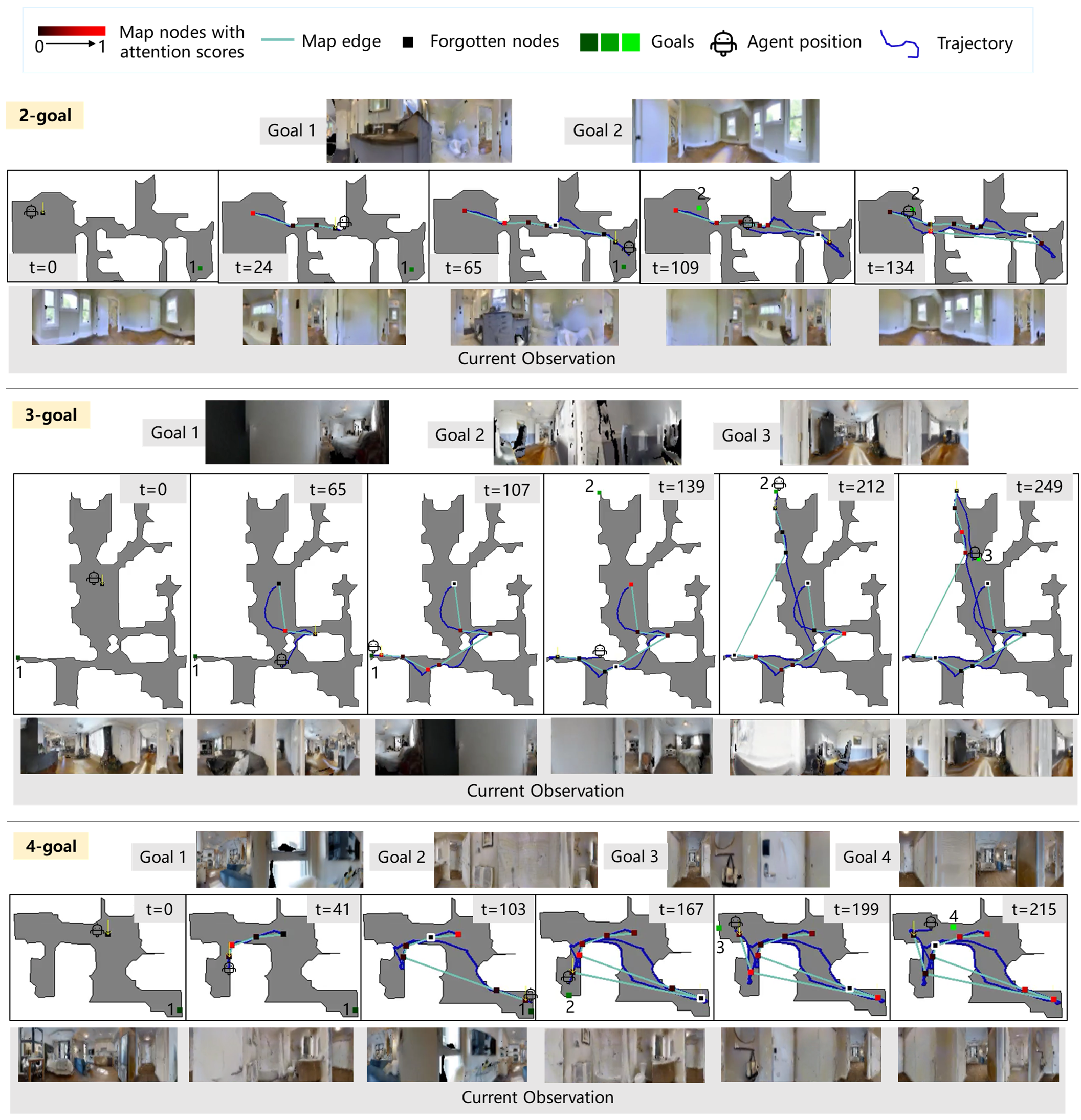}
    \caption{\textbf{Multi-goal example trajectories of MemoNav.} Each example shows both the topological map and the trajectory. The graph nodes are incrementally added to the map and selectively retained by the forgetting module in MemoNav. The examples illustrate that MemoNav flexibly neglects distant nodes. The yellow downward arrow denotes the current localized node of the agent. The comparison with VGM in these example tasks is recorded in the supplementary videos.}
    \label{fig:supp:traj_demos}
\end{figure*}

We plot example trajectories of MemoNav in multi-goal tasks, as shown in~\cref{fig:supp:traj_demos}. These examples show that MemoNav utilizes the adaptive WM to plan efficient paths in challenging multi-goal navigation. For instance, in the 3-goal trajectory, MemoNav strategically forgets the distant topmost node during the first goal and similarly neglects the goal-irrelevant bottom-left nodes in subsequent goals.

To provide a comprehensive overview, we also analyze MemoNav's limitations by examining its failure episodes in~\cref{fig:supp:failure_demos} of the appendix. These failures predominantly fall into four categories: \emph{Stopping mistakenly}, \emph{Missing the goal}, \emph{Not close enough}, and \emph{Over-exploring}.
This analysis not only helps in understanding the limitations of MemoNav but also guides potential improvements in future iterations.

\section{Conclusion}
This paper proposes MemoNav, a novel memory model for ImageNav. This model flexibly retains informative short-term navigation memory via a forgetting module. We also introduce an extra global node as long-term memory to learn a scene-level representation. The retained short-term memory and the long-term memory are encoded by a graph attention module to generate the working memory that is used for generating action. The experimental results show that the MemoNav outperforms previous methods in multi-goal tasks and plans more efficient routes.
\section{Acknowledgments}
This work was supported in part by the National Key R\&D Program of China (No. 2022ZD0160102, No. 2022ZD0160100), the National Natural Science Foundation of China (No. U21B2042, No. 62072457), and in part by the 2035 Innovation Program of CAS.

{
    \small
    \bibliographystyle{ieeenat_fullname}
    \bibliography{main}
}

\clearpage
\setcounter{page}{1}
\maketitlesupplementary

%
\section{Relation between MemoNav and Representative Working Memory Models}
\label{sec:supp:representative WM}
\begin{figure}
  \centering
  \includegraphics[scale=0.4]{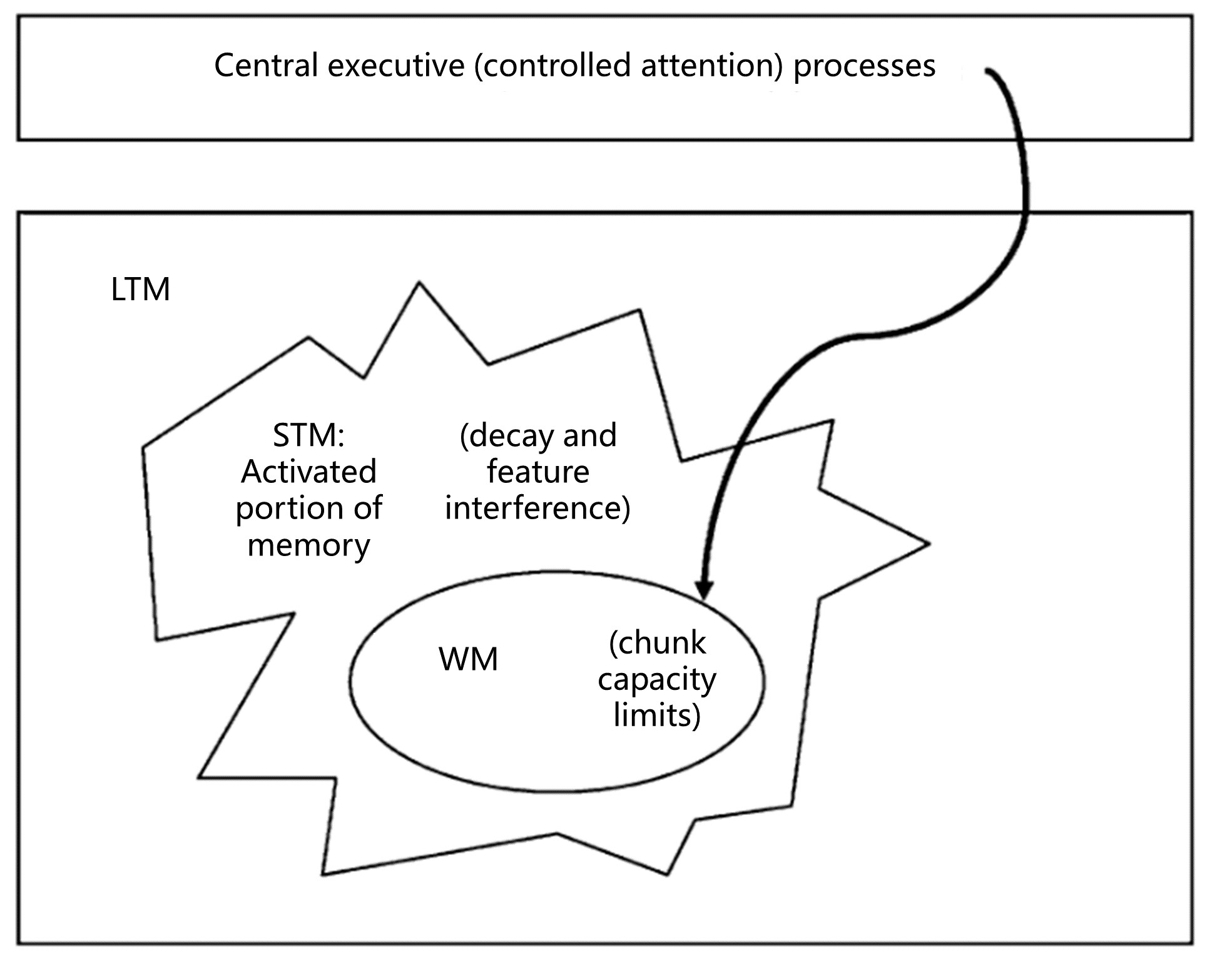}
  \caption{The memory model by Cowan et al. \citep{cowan2008differences}. This figure is borrowed and adapted from its original paper.}  
  \label{fig: cowan}
\end{figure}
Human memory consists of complex interactions between long-term memory (LTM), short-term memory (STM), and working memory~\cite{cowan2008differences}. As defined by Cowan et al.~\cite{cowan2008differences}, LTM refers to the vast, stable knowledge base and experiences stored over a lifetime. STM is a transient, limited-capacity memory system that holds information in an accessible state for brief periods. Working memory incorporates selective parts of STM as well as stored LTM knowledge through an attention mechanism, in order to actively process information relevant to the current task or decision. Cowan et al.~\cite{cowan2008differences} also designed a framework depicting how WM is formed from STM and LTM (shown in~\cref{fig: cowan}). This framework demonstrates that STM cooperates with LTM and decays as a function of time unless it is refreshed. The useful fraction of STM is incorporated into WM via an attention mechanism to avoid misleading distractions. Subsequent work by Baddeley et al.~\cite{baddeley2012working} suggests that the central executive manipulates memory by incorporating not only part of STM but also part of LTM to assist in making a decision.

We draw inspiration from the work by Cowan et al.~\cite{cowan2008differences} and Baddeley et al.~\cite{baddeley2012working} and formulate the navigation memory of MemoNav as an emulation of the human STM, LTM and working memory systems.

The parallel between MemoNav and the two relevant models above is shown in the following list:
\begin{itemize}
    \item The map node features are termed ``STM'', since they are local and transient.
    \item The topological map of MemoNav maintains a 100-node queue to store map nodes. This design simulates STM that holds a limited amount of information in a very accessible state temporarily in the human brain.
    \item MemoNav introduces a global node aggregating prior observation features stored in the topological map, thereby simulating LTM which acts as a large knowledge base.
    \item MemoNav utilizes a forgetting mechanism to remove a fraction of STM with attention scores lower than a threshold. This mechanism acts as a simple way of decaying STM.
    \item The forgetting mechanism helps WM include part of STM.
    \item MemoNav incorporates the retained STM and the LTM into WM, which is subsequently used to generate navigation actions. This design simulates the working memory model by~\cite{baddeley2012working}.
\end{itemize}

\section{Implementation Details}
\label{sec:supp:impl details}

\begin{figure*}[th]
    \centering
    \includegraphics[width=0.99\linewidth]{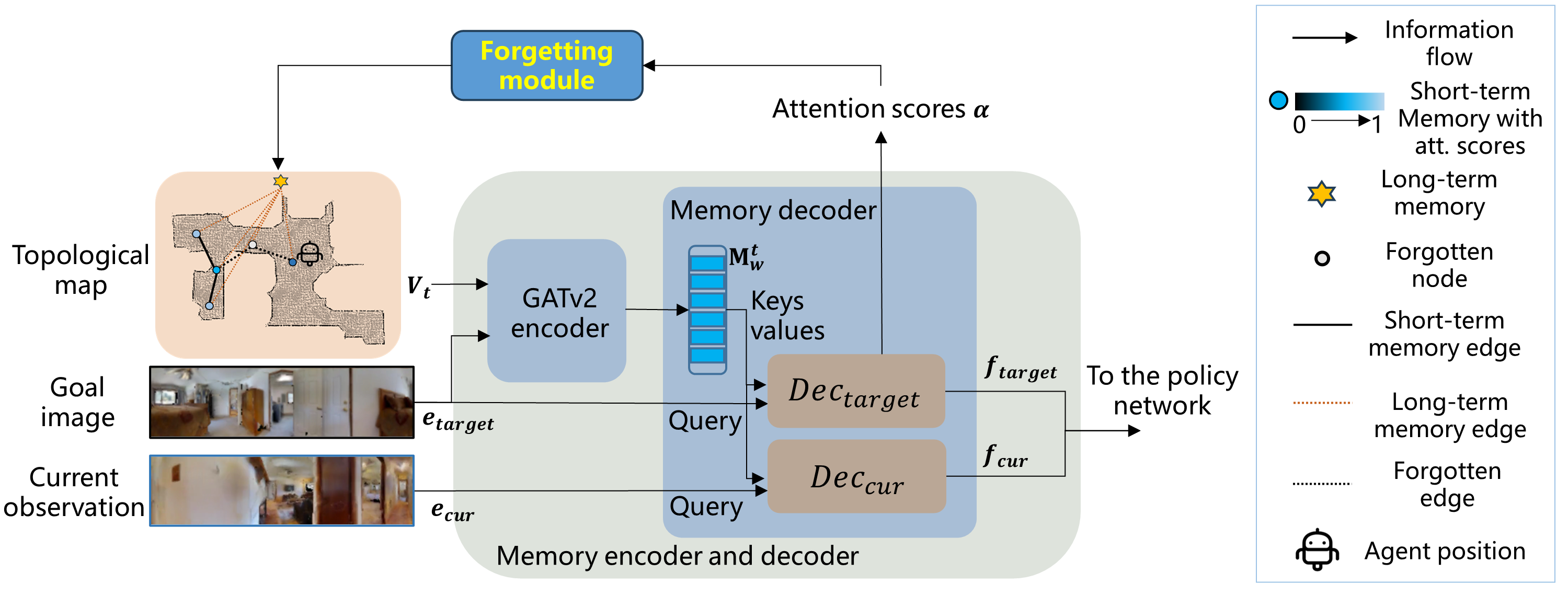}
    \caption{\textbf{The detailed structure of MemoNav.} The goal decoder $\mathcal D_{target}$ calculates the attention scores  $\mathbf{\alpha}$ for each STM feature in the topological map. Then the scores are used by the proposed forgetting module to remove redundant STM which will no longer be utilized for downstream action generation. $\displaystyle \mV$ denotes the retained STM and $\displaystyle \mM_w$ the working memory.}
    \label{fig:detailed_memonav}
\end{figure*}

\begin{algorithm*}
\label{alg: MemoNav}
\caption{The implementation of the MemoNav}\label{algorithm}
\KwData{Empty topological map $\displaystyle \gG=\{\gV,\gE\}$, goal image $\displaystyle \tI_{goal}$, current time step $\displaystyle t$, forgetting percentage $\displaystyle p$, trainable observation encoder $\mathcal F_{enc}$,  GATv2-based encoder $\operatorname{GATv2}$, Transformer decoders $\mathcal D_{goal}$ and $\mathcal D_{cur}$, LSTM-based policy network $\operatorname{LSTM}$}
\KwResult{Navigation action $\displaystyle a_{t}$}
Long-term memory $\displaystyle \vn_{global}  \leftarrow \mathbf{0} \in \mathbb{R}^d$\;
Attention scores for graph nodes $V$: $\mathbf{\alpha} \leftarrow \mathbf{0} \in \mathbb{R}^{\lvert V \rvert}$\;
\While{not $\operatorname{AgentCallStop}\left(\right)$}{
\tcp{Step 1: Update the topological map}
$\displaystyle \tI_t \leftarrow \operatorname{GetCurrentPanorama}()$\;
$\displaystyle G.\operatorname{UpdateMap}(\tI_t)$\; 

\tcp{Step 2: Retain the informative fraction of the STM}
Forgotten number $\displaystyle n \leftarrow \operatorname{Floor}\left(p \cdot \lvert \gV \rvert \right)$\;
Sorted indices $\displaystyle i \leftarrow \operatorname{Argsort}(\mathbf{\alpha})$\;

Forgotten indices $\displaystyle i_{forgotten} \leftarrow i\left[0\colon n\right]$\;
$ \displaystyle G.\operatorname{RemoveNodes}(i_{forgotten})$\;

$\displaystyle \mV \in \R^{\lvert \gV \rvert \times d} \leftarrow G.\operatorname{GetNodeFeatures}\left(\right)$\;
Working memory $\displaystyle \mM_w \leftarrow \operatorname{GATv2} (\left\{\mV, n_{global}\right\})$\ \tcp{Note that STM is fused before being forgotten in the next step so the features of forgotten STM have been fused into LTM.};
$\displaystyle \ve_{cur} \leftarrow\mathcal{F}_{enc}(\tI_{t}), \ve_{goal} \leftarrow \mathcal{F}_{enc}(\tI_{goal})$\;
$f_{cur} \leftarrow \mathcal D_{cur}\left(e_{cur},\mM_w \right),\  f_{goal} \leftarrow \mathcal D_{goal}(\ve_{goal},\mM_w )$\;
$\mathbf{\alpha} \leftarrow \mathcal D_{goal}.\operatorname{GetAttScores}()$

\tcp{Step 4: Action generation}
$\displaystyle \vx \leftarrow \operatorname{LSTM}(\operatorname{FC}(\left[\vf_{cur}, \vf_{goal}, \ve_{cur}\right]))$\;
$\displaystyle p\left(a_{t} \mid x\right)=\sigma(\operatorname{FC}(x))$\;
$\displaystyle a_{t} \leftarrow \operatorname{SampleFromDistribution}(p (a_{t} \mid \vx))$\;
}

\end{algorithm*}

\subsection{Implementation of MemoNav}
Built upon VGM, MemoNav inherits its topological map and uses its localization approach to add nodes. In addition, MemoNav improves the memory module while keeping the visual encoder and policy network unchanged of VGM.

We follow the training pipeline in~\cite{kwon2021visual} to reproduce CNNLSTM and train our MemoNav. These methods are first trained via imitation learning, minimizing the negative log-likelihood of ground-truth actions. Next, the agents are further fine-tuned with PPO~\cite{schulman2017proximal} to enhance exploratory ability. The reward setting and auxiliary losses remain the same as in VGM.The reward setting and auxiliary losses remain the same as in VGM. 

The detailed MemoNav framework is shown in~\cref{fig:detailed_memonav}. The structure of the memory decoding module in MemoNav remains the same as in VGM~\cite{kwon2021visual}. The forgetting module of MemoNav requires the attention scores generated in the decoder $\mathcal D_{goal}$. Therefore, our model needs to calculate the whole navigation pipeline before deciding which fraction of the STM should be retained. This lag means that the retained STM is incorporated into the WM at the next time step. The pseudo-code of MemoNav is shown in Algorithm 1.

\subsection{Reproduction of CNNLSTM}
We reproduce CNNLSTM~\cite{zhu2017target} following the description in its original paper, but we also make some modifications to keep the comparison fair. We replace the ResNet-50 in CNNLSTM with the pretrained RGB-D encoder of VGM~\cite{kwon2021visual}. We also add positional embeddings to the encoded RGB-D observations to contain temporal information. Moreover, we concatenate the encoded RGB-D observations with the goal image embedding and project the concatenated feature (1024D) to a 512D feature, so CNNLSTM can utilize the information of the goal image. The projected features of four consecutive frames are further condensed and then input to a policy network as in~\cite{zhu2017target}. To use the two auxiliary tasks proposed in VGM~\cite{kwon2021visual}, we also introduce the linear projection layers (Linear-ReLU-Linear) used in VGM to process the embedded goal image and embedded current observation.

\subsection{Compute Requirements}
We utilize an RTX TITAN GPU for training and evaluating our models. The imitation learning phase takes 1.5 days to train while the reinforcement learning takes 5 days.

The computation in the GATv2-based encoder and the two Transformer decoders occupy the largest proportion of the run-time of MemoNav. The computation complexity of the encoder and the decoders are $\displaystyle \mathcal{O}(\lvert \gV \rvert d^2 + \lvert \gE \rvert d)$ and $\displaystyle \mathcal{O} \left(\lvert \gV \rvert d\right)$, respectively. Using the forgetting module with a percentage threshold $p$, the computation complexity of MemoNav can be flexibly decreased by reducing the number of nodes to $(1-\displaystyle p)\lvert \gV \rvert$.

\section{Comparison between MemoNav with and without Forgetting}
\label{sec:supp:comp wo forgetting}
\begin{figure*}[th]
    \centering
    \includegraphics[width=0.99\linewidth]{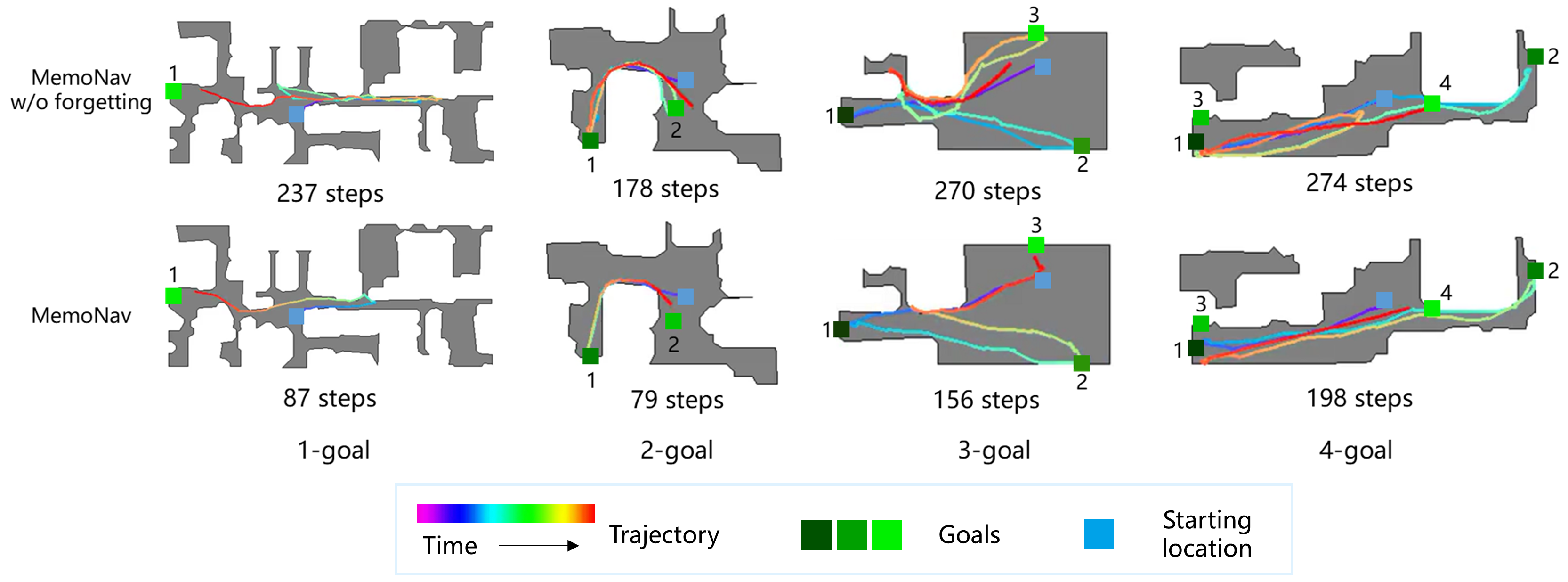}
    \caption{\textbf{Visualization comparing the MemoNav with and without the forgetting module.} We compare selected episodes at four difficulty levels in the Gibson scenes and visualize the top-down views. MemoNav without the forgetting module exhibits more sharp turns and tends to take more steps, demonstrating lower efficiency compared to the full MemoNav. The number of navigation steps (the upper limit is 500) are shown at the bottom of each top-down view. Best viewed in color.}
    \label{fig: ablation vis}
\end{figure*}

We analyze the impact of the forgetting module on MemoNav's trajectory properties, such as smoothness and length. \cref{fig: ablation vis} illustrates that the inclusion of the forgetting module results in more smooth and efficient trajectories. In contrast, trajectories generated without this module are characterized by numerous abrupt turns and extended paths. This disparity likely arises from a segment of the Short-Term Memory (STM) containing irrelevant information, leading to frequent and erratic alterations in the policy network's action output. The forgetting module effectively filters out this disruptive portion of STM, thereby enabling the policy network to use task-relevant navigation memory for efficient decision-making.

\section{In-depth Analysis of Forgetting Module}
\label{sec:supp:analysis-forget}

\begin{figure*}[th]
    \centering
    \includegraphics[width=0.8\linewidth]{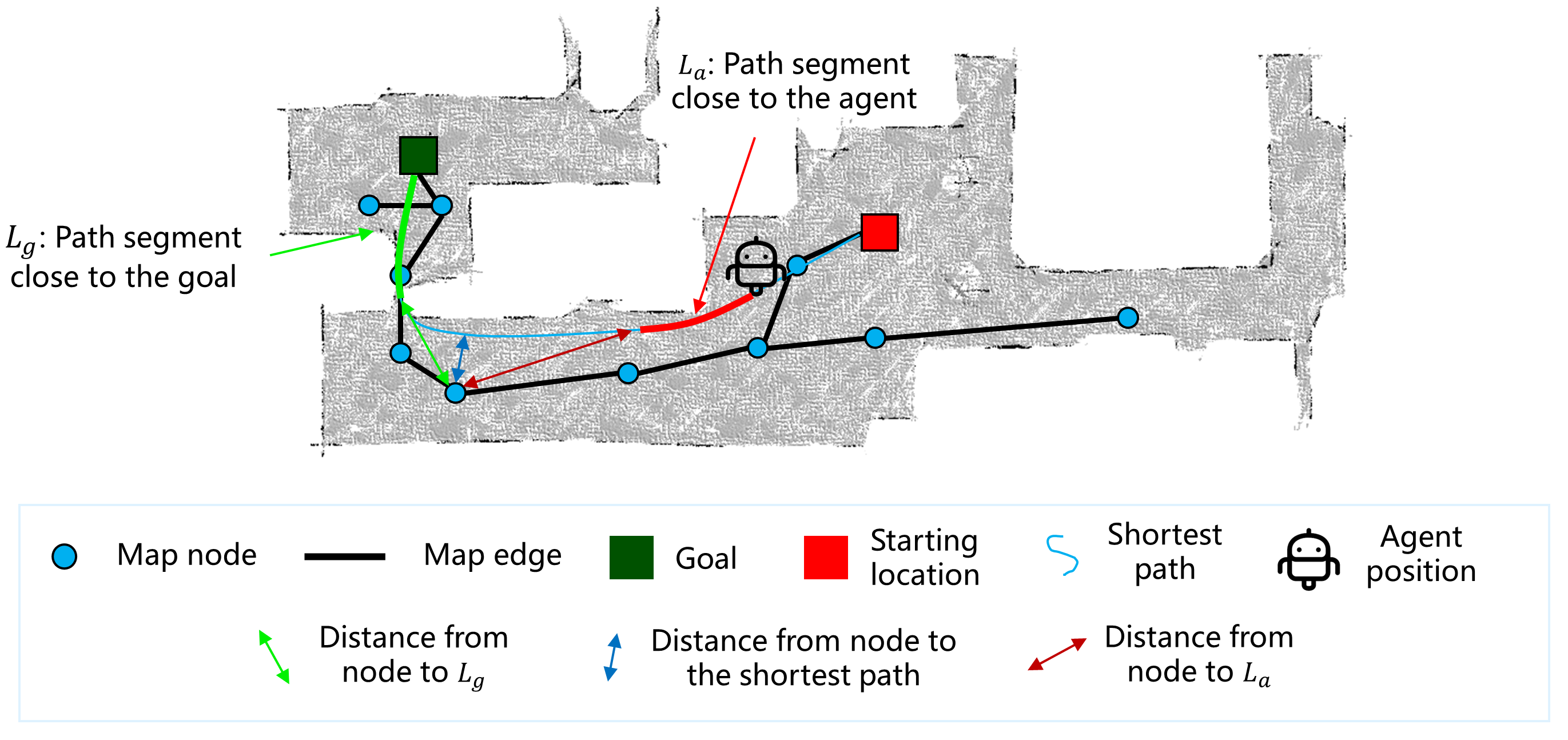}
    \caption{The visualization of distance metrics (c), (d), and (e) defined in~\cref{sec:supp:analysis-forget}.}
    \label{fig:dist metrics}
\end{figure*}

\begin{figure*}
  \centering
  \includegraphics[width=0.8\linewidth]{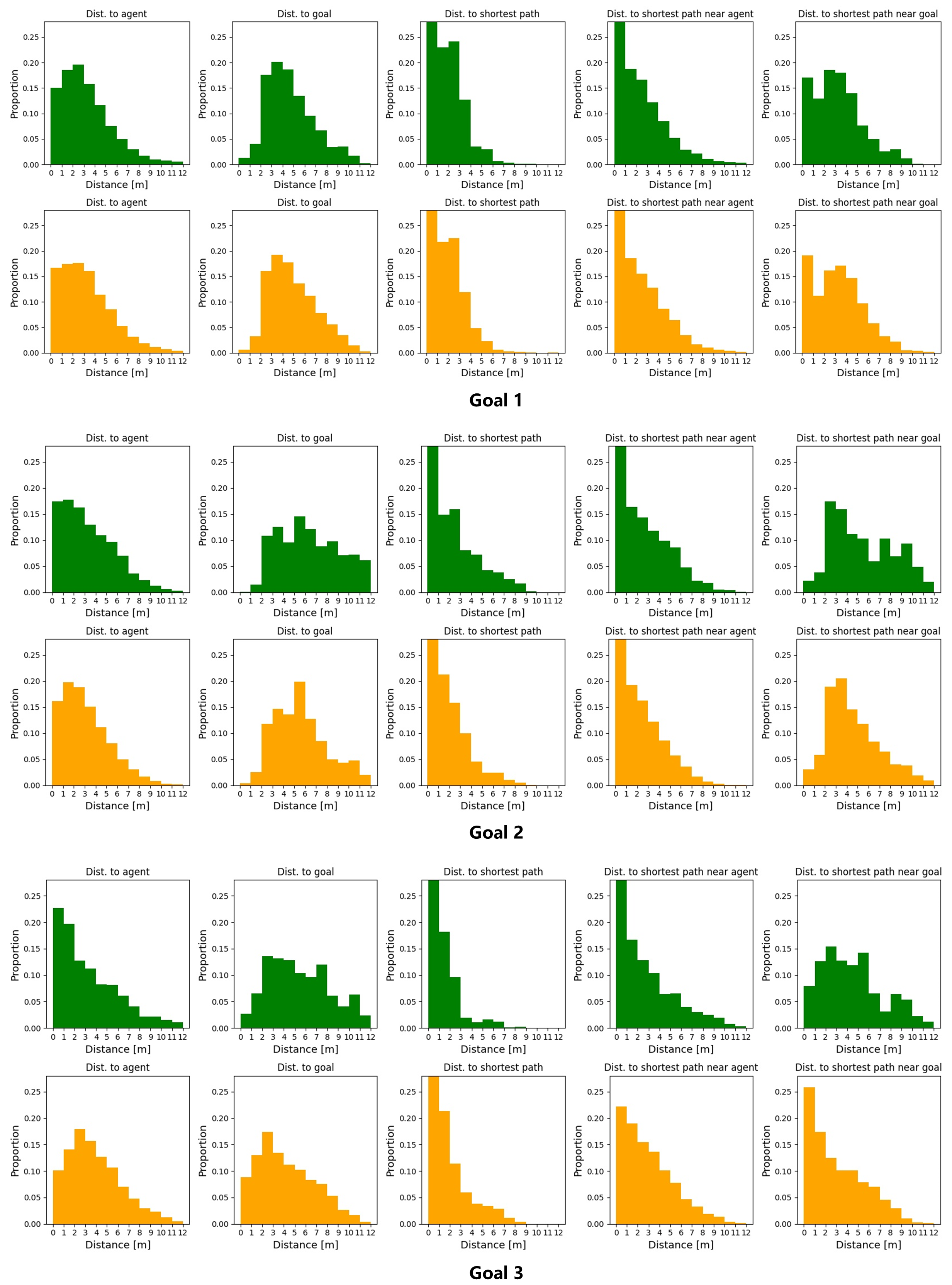}
  \caption{\textbf{Histograms of the five distance metrics defined in~\cref{sec:supp:analysis-forget}}. The data of these metrics is collected by evaluating the MemoNav on the 3-goal task in the Gibson scenes and averaged over five runs. The upper row (green) and lower row (orange) belong to the forgotten nodes and retained ones, respectively.}  
  \label{fig:supp:disthist}
\end{figure*}

An extensive statistical analysis is conducted to comprehend the forgetting module's functionality. In this experiment, five distance metrics are calculated: (a) distance from a node to the \textbf{agent}, (b) distance from a node to the \textbf{goal}, (c) distance from a node to \textbf{the oracle shortest path}, (d) distance from a node to the shortest path segments closer to the \textbf{agent}, and (e) distance from a node to the shortest path segments closer to the \textbf{current goal}. Then the histograms of these five metrics are drawn according to the metrics records for each forgotten/retained node at each time step so we can see the patterns of these distance metrics. Please see~\cref{fig:dist metrics} to better understand the definitions of the distance metrics (c)(d)(e).

We evaluate MemoNav on the 3-goal Gibson task and draw the histograms \textbf{on per-goal basis}, as shown in~\cref{fig:supp:disthist}. The figure provides two interesting findings:

\begin{itemize}
\item The distance distribution patterns for forgotten nodes (green bars) and retained ones (orange bars) vary across goals. Notably, as the agent progresses to the third goal, the distributions of the distances from forgotten nodes to goals (column 2) and to shortest path segments near goal (column 5) become uniform. In contrast, these two histograms for the retained nodes become sharper and the peaks shift to smaller distance values. This pattern suggests the forgetting module selectively retain nodes that are proximal and relevant to the current goal.
\item The forgetting module has a larger impact on the distance metrics when the navigation task becomes more difficult. Specifically, when the current goal index is 1 (i.e. the task is easy), the averages of the distance metrics for forgotten nodes and retained nodes are close. When the goal index rises to 3 (i.e. the task becomes harder), a larger proportion of the retained nodes are close to the goal, the shortest path, and the shortest path segments near goal. This pattern suggests that MemoNav focuses on critical areas for navigation, such as the goal vicinity and the shortest path.
\end{itemize}

These results empirically validate that MemoNav is able to retain the information useful for multi-goal navigation via the forgetting module.

\begin{figure*}[th]
    \centering
    \includegraphics[width=0.95\linewidth]{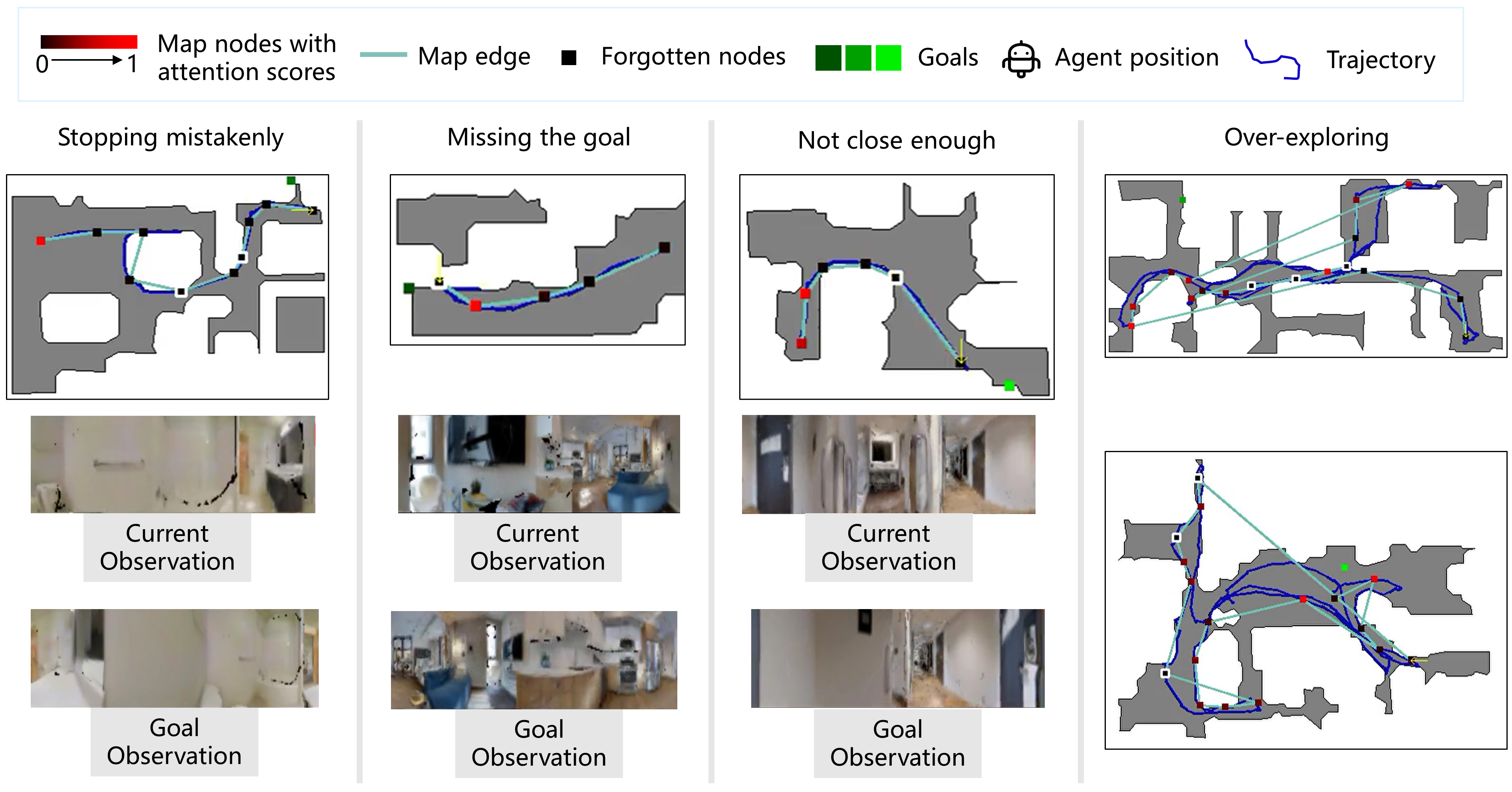}
    \caption{\textbf{Examples of failed episodes.} The agent encounters four major failure mode: (1) \emph{Stopping mistakenly}: the agent implements stop at the wrong place. (2) \emph{Missing the goal}: the agent has observed the goal but passes it. (3) \emph{Not close enough}: the agent attempts to reach the goal it sees but implements stop outside the successful range. (4) \emph{Over-exploring}: the agent spends too much time exploring open areas without any goals.}
    \label{fig:supp:failure_demos}
\end{figure*}

\section{The Variation of the LTM}
\begin{figure*}
  \centering
  \includegraphics[width=0.9\linewidth]{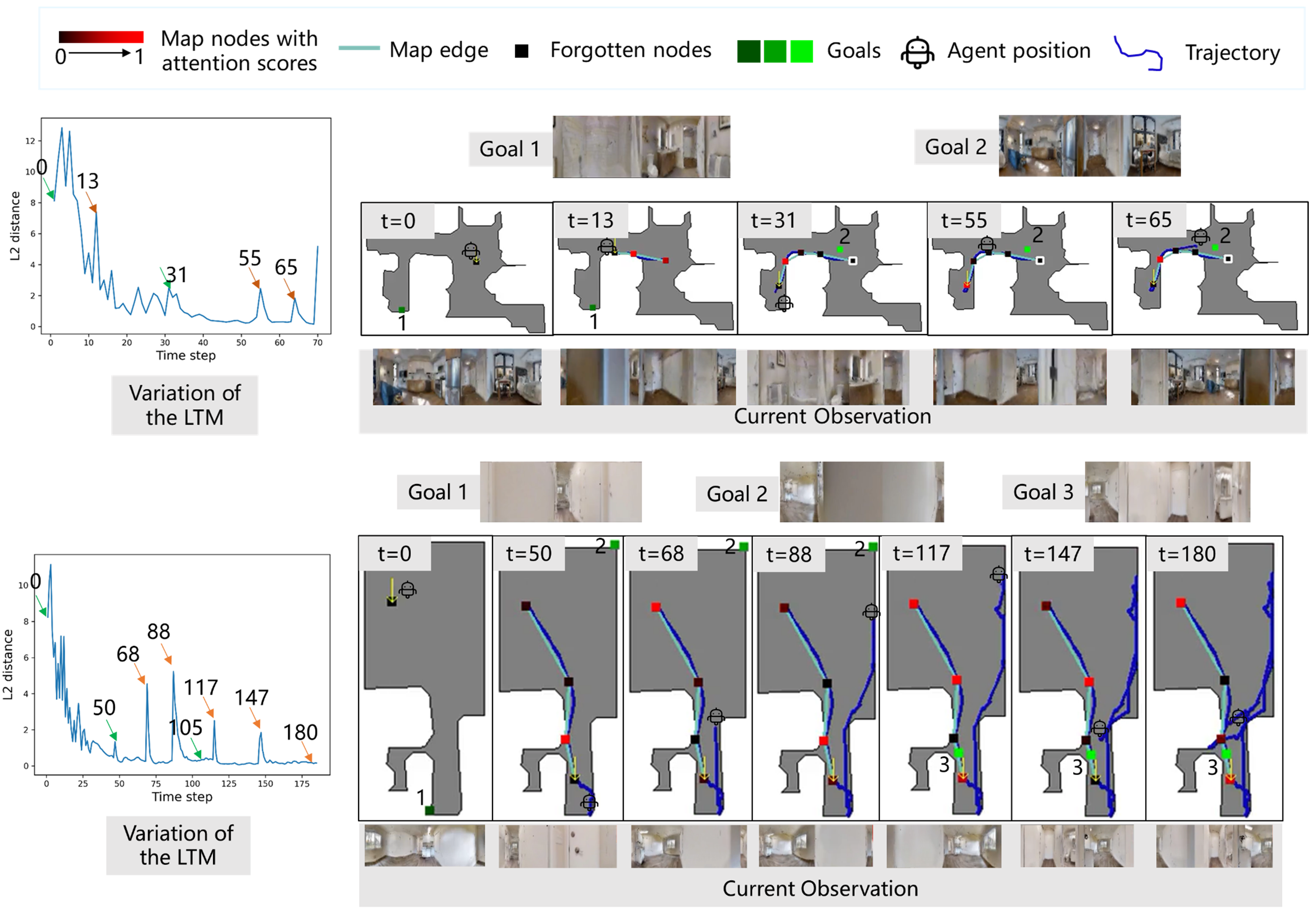}
  \caption{\textbf{Visualization of the LTM variation.} We show the agent's trajectories in two example episodes and visualize the agent's observations at the time steps when peaks appear on the LTM variation curves. The green arrows denote when the agent sets a new goal while the orange ones denote when peaks appear.}  
  \label{fig:LTM_variation}
\end{figure*}
We explore the dynamic nature of the LTM during navigation by calculating the L2 distance between consecutive time-step features, as depicted in~\cref{fig:LTM_variation}. The trends observed in these curves – rapid initial increases in L2 difference followed by stabilization and intermittent peaks – are indicative of the LTM's response to the agent's environmental interactions.

To understand why the LTM variation shows such a trend, we visualize the agent's observations at the time steps of the peaks. Specifically, the L2 difference remains low in familiar areas, suggesting stability in the LTM's feature representation. For instance, in the 2-goal example (top row), the L2 difference steadily decreases in $t=31\sim 55$ during which the agent travels around visited areas; (2) The L2 difference increases sharply upon encountering new scenes. These peaks correspond with the agent's exposure to novel views. For instance, in the 3-goal example (bottom row), the L2 difference curve exhibits peaks at $t=68$ when the agent passes a corner and at $t=88$ when the agent observes a novel open area. These results highlighting the LTM's role in assimilating new exploratory experiences.

\section{Limitations}
While MemoNav witnesses a large improvement in the navigation success rate in multi-goal navigation tasks, it still encounters limitations. The proposed forgetting module is a post-processing method, as it obtains the attention scores of the decoder before deciding which nodes are to be forgotten. Future work can explore trainable forgetting modules. The second limitation is that our forgetting module does not reduce memory footprint, since the features of the forgotten nodes still exist in the map for localization. Moreover, the forgetting threshold in our experiments is fixed. Future work can merge our idea with Expire-span \citep{sukhbaatar2021not} to learn an adaptive forgetting threshold.

\section{Potential Impact}
The notable potential of negative societal impact from this work: our model is trained on 3D scans of the Gibson scenes which only contain western styles. This inadequacy of diverse scene styles may render our model biased and incompatible with indoor environments in unseen styles. As a result, our model may be only available in a small fraction of real-life scenes. If our model is transferred to out-of-distribution scenes, the agent may take more steps and even bump on walls frequently.

\end{document}